\documentclass{article}
\usepackage{iclr2027_conference,times}
\usepackage[T1]{fontenc}
\usepackage{amsmath}
\usepackage{amssymb}
\usepackage{graphicx}
\usepackage[hidelinks]{hyperref}
\usepackage{url}
\title{Halo: Improving forecast accuracy through heteroscedastic estimation}
\author{Adam Cataldo \\
AI Quant Researcher \\
\texttt{adam.cataldo@gmail.com}}
\iclrfinalcopy
\begin{document}
\maketitle
\lhead{}
\renewcommand{\headrulewidth}{0pt}

\begin{abstract}
Heteroscedastic forecasting, where a network estimates a scale parameter alongside a location parameter, is normally motivated by uncertainty quantification. This paper shows it also improves the point estimate, in contrast to reported negative results for heteroscedastic estimation outside time series. Halo is a modification that reuses an existing deep forecaster's architecture, giving it a second output for the scale of its implied distribution and training it under the matching negative log likelihood. Adapting three state-of-the-art models --- a transformer, a graph network paired with a variational autoencoder, and a single-layer convolutional network --- under both Gaussian and Laplacian losses demonstrates the phenomenon. On the five electricity price markets of a standard forecasting benchmark, Halo improves MSE and MAE in 28 of 30 model-market-metric comparisons, cutting average MSE by 2.6\% to 16.5\% and average MAE by 1.7\% to 11.0\%. Two findings emerge: (1) whether the scale estimate comes from a second projection head or from a full parallel network matters far less than whether the network estimates scale, and (2) the improvement holds under the hyperparameters already tuned for the point-estimate baseline, so retuning is optional.
\end{abstract}

\section{Introduction}
\label{sec:introduction}

Time series forecasting with deep neural networks (DNNs) poses some unique challenges relative to other DNN problems, like computer vision and natural language processing. On the surface, it shares similarities with language models, where the input is a sequence of words (represented as tokens) and the output is the next sequence of words that follows the input. In time series, the input is a sequence of data points, and the output is the next sequence of data points. Unlike language modeling, however, the data points themselves carry no semantic meaning the way words do \citep{nie2022}. For many real-world time series, the past values of the time series itself are less informative about the future values than past values of related (exogenous) time series \citep{granger1969,box2015}. For example, in forecasting stock price returns, past values of the stock price tend to have less influence on future returns than do exogenous time series like interest rates and company fundamentals \citep{chincarini2022}. There are other challenges as well: since time series are rarely stationary, the older data a model trains on may have a different distribution than the newer data it runs inference on \citep{kim2021,liu2022}. Additionally, model training has to avoid leakage, where data from the test or validation set influences the trained model and gives false confidence in its forecasting ability. While obvious forms of leakage are easy to stop, more subtle forms can go unnoticed and skew results, like an exogenous time series sampled in the future, relative to the endogenous time series being predicted \citep{hewamalage2023}.

In this paper, we introduce a technique for modifying an existing time-series deep learning architecture to improve accuracy. The basic idea is to use heteroscedastic forecasting, where the forecaster learns scale parameters for the output, in addition to location parameters. Making heteroscedastic forecasts is useful by itself, since it helps quantify the uncertainty of a forecast. We make a stronger claim, which is that learning the scale parameter helps the forecaster improve its location estimate. We call this technique \emph{Halo}, to emphasize that finding the halo (distribution) around the point estimate makes it easier for the network to locate the point estimate itself. This paper explores the subtleties of the Halo method, like how to architect the location and scale estimates, and how heteroscedastic estimation interacts with nonstationarity adjustments. The result is a technique that boosts performance on several state-of-the-art (SOTA) models.

\section{Related Work}
\label{sec:related-work}

Neural networks have been used for time series forecasting since at least the 1980s \citep{lapedes1987,werbos1988}. In the 2010s, deep neural networks became popular, with recurrent neural network (RNN) approaches like DA-RNN \citep{qin2017}, LSTNet \citep{lai2018}, and ES-RNN \citep{smyl2020}. TCN \citep{borovykh2017} showed that convolutional neural networks (CNNs) could outperform RNNs for forecasting tasks, and N-BEATS \citep{oreshkin2019} showed how to build effective DNNs without using RNNs or CNNs. Like other areas of deep learning, transformers \citep{vaswani2017} have become popular in time series forecasting. Some early transformer models like FEDFormer \citep{zhou2022}, Autoformer \citep{wu2021}, and Informer \citep{zhou2021} were shown to be less effective than simple linear models for forecasting tasks \citep{zeng2023}. PatchTST \citep{nie2022} showed that tokenizing patches of time series (subsequences), rather than tokenizing individual data points in a series, led to more meaningful tokens, and effectively unlocked transformers as a valuable tool for time series forecasting. TimeXer \citep{wang2024} effectively combined patching with exogenous time series for more general-purpose time series forecasting. GCGNet \citep{li2026} was able to combine patching, a graph neural network (GNN), and a variational autoencoder (VAE) to outperform TimeXer. By default, GCGNet has access to future exogenous inputs, exactly the type of leakage that \citet{hewamalage2023} argues against. Nonetheless, the authors showed that even when the future exogenous inputs are not available, it can outperform TimeXer, though its performance is degraded. \citet{zhou2025} created a competitive model, CrossLinear, from patching and a single-layer CNN alone.

Heteroscedastic estimation in neural networks dates back to the 1990s \citep{nix1994}. While point estimates are most common in the time-series deep learning literature, DeepAR \citep{salinas2020} does take a probabilistic modeling approach. Interestingly, \citet{stirn2023} showed a result that on the surface seems to contradict the main claim of this paper: on four neural network models, heteroscedastic estimation led to worse point estimation. None of these models were time series forecasters, and all were much smaller in scale than the models considered in this paper. This suggests potential future work identifying exactly when heteroscedastic estimation does and does not improve point estimates, since it does not seem to be a universal result in neural networks.

\section{Halo}
\label{sec:halo}

The basic forecasting problem this paper works through is as follows. A network takes inputs $(x_{endo}, x_{exo}) \in \mathbb{R}^T \times \mathbb{R}^{T \times C}$ and generates a forecast $\hat{y} \in \mathbb{R}^{H}$. $x_{endo}$ represents the last $T$ data points for the \emph{endogenous signal} to be forecast. $\hat{y}$ is an estimate of $y \in \mathbb{R}^{H}$, the next $H$ data points in the signal. Letting $s$ represent the underlying endogenous signal of interest, at time $t$, $(x_{endo}(1), \ldots, x_{endo}(T)) = (s(t-T+1), \ldots, s(t))$ and $(y(1), \ldots, y(H)) = (s(t+1), \ldots, s(t+H))$. $T$ is the \emph{lookback window} of the forecast, and $H$ is the \emph{forecast horizon}. $x_{exo}$ represents the last $T$ data points of $C$ \emph{exogenous inputs}, signals related to the endogenous signal, which can help predict its future values. $C$ is the number of \emph{exogenous channels}.

The network is a function $f: \mathbb{R}^T \times \mathbb{R}^{T \times C} \times \Phi \to O$, where $\Phi$ is the space of network parameters and $O$ is the output space. The network learns its parameters during training and holds them constant during inference. For point estimates, $O = \mathbb{R}^{H}$, and $\hat{y} = f(x_{endo}, x_{exo}, \phi)$. For heteroscedastic estimates, $O = \mathbb{R}^{H} \times (0, \infty)^H$. If the distribution parameters for location and scale are $a$ and $b$ respectively, write $(\hat{a}, \hat{b}) = f(x_{endo}, x_{exo}, \phi)$. For example, for a Gaussian distribution with mean $\mu$ and variance $\sigma^2$, $(\hat{\mu}, \hat{\sigma}^2) = f(x_{endo}, x_{exo}, \phi)$.

Simplifying assumptions:
\begin{itemize}
\item Assume the endogenous signal is a single channel. This is for clarity, and not a fundamental limitation of heteroscedastic estimation.
\item Assume each exogenous signal has the same lookback window as the endogenous signal. Again, this is to keep exposition simple, and not a fundamental limit of the approach.
\item Assume future exogenous inputs are not visible to the forecaster. Some recent architectures like GCGNet \citep{li2026} and DAG \citep{qiu2025} have let the forecaster see future exogenous inputs, under the view that some exogenous inputs can have highly accurate forecasts. This paper avoids that, in the interest of limiting leakage \citep{hewamalage2023}, since in general future exogenous inputs may carry information about yet-to-be-predicted endogenous values. Also, a forecaster can treat predictions of future values made at time $t$, about time $t+h$, as exogenous values at time $t$, so this is not a fundamental limitation.
\item Assume the distribution used for heteroscedastic loss is parameterized by a location and scale parameter $a$ and $b$. That is the case for the examples considered in this paper, but not a requirement for heteroscedastic loss. For example, a Weibull distribution has three parameters and an exponential distribution has only a single parameter.
\end{itemize}

\subsection{Heteroscedastic loss}
\label{sec:heteroscedastic-loss}

This subsection explores several loss functions of the form $L\left((y_1, \ldots, y_N), (\hat{y}_1, \ldots, \hat{y}_N)\right)$, where each $y_i$ and $\hat{y}_i$ are members of $\mathbb{R}^{H}$. That represents a loss over $N$ samples and a forecast horizon $H$. To keep the exposition clear, a loss function written like:

\begin{equation}
\frac{1}{N} \sum_{i = 1}^N (y_i - \hat{y}_i)^2,
\end{equation}

is really shorthand for

\begin{equation}
\frac{1}{H} \sum_{t=1}^H \left[ \frac{1}{N} \sum_{i = 1}^N (y_i(t) - \hat{y}_i(t))^2 \right].
\end{equation}

That is to say that in practice, the loss is averaged across the forecast horizon, but to keep equations short and readable, they are written in this subsection as if the forecast horizon was 1, and there was only a single output value to consider.

The two most common loss functions in deep learning time-series literature are mean squared error (MSE) and mean absolute error (MAE). These are defined by

\begin{align}
\text{MSE} &= \frac{1}{N} \sum_{i = 1}^N (y_i - \hat{y}_i)^2, \\
\text{MAE} &= \frac{1}{N} \sum_{i = 1}^N |y_i - \hat{y}_i|.
\end{align}

Using either of these as loss functions for training has a probabilistic interpretation. Minimizing either loss function corresponds to minimizing the negative log likelihood (NLL) function for some assumed distribution $P(Y \mid X)$, where $Y$ is the output random variable and $X$ is the input random variable. More specifically, given learned model parameters $\phi$, write $x_i$ for the input pair $(x_{endo}, x_{exo})$ of sample $i$, and treat the estimate $\hat{y}_i = f(x_i, \phi)$ of $y_i$ as the location parameter for some implied distribution function. Here $f$ is the network function.

If this distribution is Gaussian $\mathcal{N}(\mu, \sigma^2)$, with some mean $\mu$ that depends on the input, and some constant, but unknown, variance $\sigma^2$, then the NLL given $N$ data points is

\begin{equation}
\label{eq:gaussian-nll}
\text{NLL} = \frac{N}{2} \log(2 \pi) + \frac{1}{2} \sum_{i=1}^N \log(\sigma^2) + \sum_{i=1}^N \frac{(y_i - \mu_i)^2}{2 \sigma^2}.
\end{equation}

In this case, with $\hat{\mu}_i = \hat{y}_i = f(x_i, \phi)$ and $\sigma^2$ constant, the $\phi$ that minimizes the MSE is the $\phi$ that minimizes the NLL.

Similarly, if the distribution is Laplacian $\mathcal{L}(\mu, b)$ with location $\mu$ and constant, but unknown, scale $b$, then minimizing the MAE is equivalent to minimizing the NLL

\begin{equation}
\text{NLL} = N \log(2) + \sum_{i=1}^N \log(b) + \sum_{i=1}^N \frac{|y_i - \mu_i|}{b}.
\end{equation}

Removing the assumption that the variance/scale parameter is constant, and modifying the network so that instead of estimating $\hat{\mu}_i = f(x_i, \phi)$ it estimates both location and scale, like $(\hat{\mu}_i, \hat{\sigma}_i^2) = f(x_i, \phi)$ in the Gaussian case, gives a heteroscedastic model. Heteroscedastic models can be useful in their own right. For instance, a heteroscedastic model can generate probabilistic forecasts, giving ranges of possible values rather than a single one. This paper provides evidence that heteroscedastic estimates are useful even when only the point estimate matters, by helping the network make better point estimates of the location parameter. Empirically, when the network learns both location and scale, it adjusts its point estimate to account for variability. That is, the network learns to account for variance.

Note that for the Gaussian distribution, this paper uses the $\beta$-NLL function introduced by \citet{seitzer2022} to find the mean and variance parameters:

\begin{equation}
\beta\text{-NLL} = \frac{1}{2N} \sum_{i=1}^N \text{sg}\left(|\sigma_i|\right) \log(\sigma_i^2) + \frac{1}{2N} \sum_{i=1}^N \text{sg}\left(|\sigma_i|\right) \frac{(y_i - \mu_i)^2}{\sigma_i^2}.
\end{equation}

Here $\text{sg}(\cdot)$ is the stop-gradient operator. It acts as an identity function in forward-pass operations, but stops gradients from flowing in backward-pass operations. The stop gradient ensures that minimizing this function is equivalent to minimizing the NLL, and \citet{seitzer2022} showed that it is more stable during backpropagation updates than the traditional Gaussian NLL. It is equivalent because it has the same optimal point, even though it has different gradients. There, the stop-gradient term carries an exponent $\beta$, which the authors recommend setting to 0.5. No $\beta$ term appears in the equation because it is already fixed at 0.5, which is what leaves $\text{sg}(|\sigma_i|)$ in its place. No such modification is needed for the Laplacian distribution.

One final note: choosing the loss function to minimize is ultimately making a choice about the assumed distribution $P(Y \mid X)$. The right choice depends on the underlying data. If the right choice is unclear, the loss function itself can be treated as a hyperparameter. In this case, the NLL function applied to the validation set can identify which distribution better fits the data. It needs to be the full NLL function, not a function that produces the same minimum, like MSE or $\beta$-NLL in place of the Gaussian NLL in Equation~\ref{eq:gaussian-nll}. Otherwise, the comparison risks a scale mismatch across distributions.

\subsection{From point estimates to heteroscedastic estimates}
\label{sec:from-point-estimates-to-heteroscedastic-estimates}

\begin{figure}[h]
\begin{center}
\includegraphics{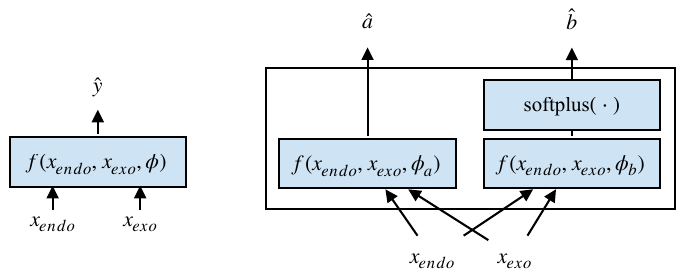}
\end{center}
\caption{Running the point-estimate network on the left twice in parallel produces the heteroscedastic network on the right.}
\label{fig:parallel}
\end{figure}

One way to go from point estimates to heteroscedastic estimates is to run two copies of the same basic network architecture in parallel, one for the location parameter and one for the scale parameter. See Figure~\ref{fig:parallel}. Note that the $\textrm{softplus}: \mathbb{R}^H \to (0, \infty)^H$ operator ensures all scale estimates are positive, with $\textrm{softplus}(c) = \ln(1 + e^c)$. While this is viable, it doubles the network size, and can waste network parameters effectively relearning the same representation twice.

\begin{figure}[h]
\begin{center}
\includegraphics{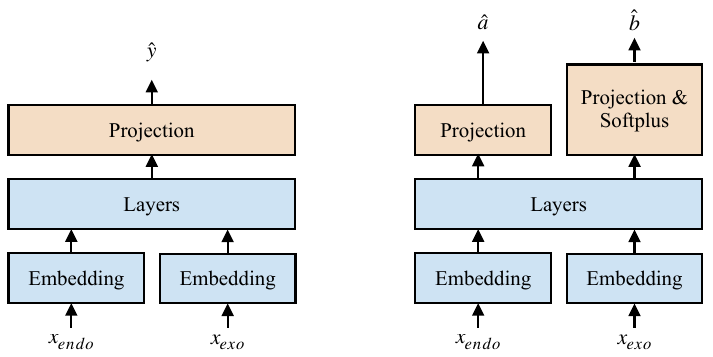}
\end{center}
\caption{Giving the point-estimate network on the left a second projection head produces the heteroscedastic network on the right.}
\label{fig:dual-head}
\end{figure}

A common architecture feeds $x_{endo}$ and $x_{exo}$ through embedding functions, then transforms them into an internal representation with multiple layers of the core architecture, and finally applies a projection to map the representation onto outputs. In this case, creating two projections is often sufficient to achieve an effective heteroscedastic model using a single learned representation of the inputs and distinct learned projection functions. See Figure~\ref{fig:dual-head}. In practice, both architectures generally outperform the non-heteroscedastic architecture. Which of the two makes better estimates is close to even, so if efficiency is a constraint, the dual-head architecture is safe, and if not, hyperparameter tuning can test both architectures as variants.

\subsection{Nonstationarity adjustments}
\label{sec:nonstationarity-adjustments}

\citet{liu2022} introduced Series Standardization as a mechanism to mitigate some of the effects of nonstationarity. The idea is to ``standardize'' the input $x_{endo}$ with:

\begin{align}
m &= \frac{1}{T} \sum_{t=1}^T x_{endo}(t) \\
v &= \sqrt{\frac{1}{T} \sum_{t=1}^T (x_{endo}(t) - m)^2} \\
x'_{endo} &= \frac{1}{v} (x_{endo} - m)
\end{align}

The network then takes $x'_{endo}$, and a final step transforms its output $\hat{y}'$ back with:

\begin{equation}
\hat{y} = \hat{y}' v + m
\end{equation}

All the models in this paper use Series Standardization. This conflicts with the Halo method as described above, because the input standard deviation $v$ has to adjust the scale estimate too. Thus, if $\hat{a}'$ is the location estimate the heteroscedastic network emits, and $\hat{b}'$ is the scale parameter after softplus, then the final outputs are:

\begin{align}
\hat{a} &= \hat{a}' v + m \\
\hat{b} &= \hat{b}' v
\end{align}

The experiments below use these adjusted outputs. The models also apply a nonstationarity adjustment to the exogenous inputs, but the specific adjustment varies across the models, and it does not interact with Halo in any meaningful way.

\section{Experiments}
\label{sec:experiments}

Three SOTA models adapted with Halo demonstrate its ability to boost forecast accuracy: TimeXer \citep{wang2024}, GCGNet \citep{li2026}, and CrossLinear \citep{zhou2025}. TimeXer and GCGNet are both complex, multi-layer models adapted with the dual-head architecture described above, whereas CrossLinear is a single-layer CNN adapted by running two networks in parallel. TimeXer is a transformer-based architecture, while GCGNet uses a GNN combined with a VAE for its architecture. The authors of TimeXer and CrossLinear trained their models with MSE loss, while the GCGNet authors trained their models with MAE loss, so these three examples enable testing the Halo method with both Gaussian and Laplacian distributions.

The datasets for testing are the electricity price forecasting (EPF) benchmarks described in detail by \citet{lago2021}. This is a set of five datasets with different exogenous and endogenous signals targeting electricity prices across different markets: Nord Pool (NP), the Pennsylvania-New Jersey-Maryland interconnection (PJM), and the Belgian (BE), French (FR), and German (DE) markets of EPEX SPOT. The datasets themselves are diverse, because of the different signals and different pricing models they track. More importantly, the original authors of TimeXer, GCGNet, and CrossLinear have already identified a good set of hyperparameters on those datasets, with tuned hyperparameters reused here for the baselines. First, the headline results: as stated above, Halo generally boosts performance across the five markets. See Table~\ref{tab:headlines}.

\begin{table}[h]
\caption{Each of the three models is compared side-by-side with and without Halo applied. Halo generally boosts performance, with the only exception being a single market where TimeXer does better without Halo. Here ``Avg'' represents the average MSE/MAE for each model across all five markets.}
\label{tab:headlines}
\begin{center}
\resizebox{\textwidth}{!}{
\begin{tabular}{|l|r|r|r|r|r|r|r|r|r|r|r|r|}
\hline
 & \multicolumn{2}{c|}{\textbf{TimeXer + Halo}} & \multicolumn{2}{c|}{\textbf{TimeXer}} & \multicolumn{2}{c|}{\textbf{GCGNet + Halo}} & \multicolumn{2}{c|}{\textbf{GCGNet}} & \multicolumn{2}{c|}{\textbf{CrossLinear + Halo}} & \multicolumn{2}{c|}{\textbf{CrossLinear}} \\
\cline{2-13}
 & \multicolumn{1}{c|}{\textbf{MSE}} & \multicolumn{1}{c|}{\textbf{MAE}} & \multicolumn{1}{c|}{\textbf{MSE}} & \multicolumn{1}{c|}{\textbf{MAE}} & \multicolumn{1}{c|}{\textbf{MSE}} & \multicolumn{1}{c|}{\textbf{MAE}} & \multicolumn{1}{c|}{\textbf{MSE}} & \multicolumn{1}{c|}{\textbf{MAE}} & \multicolumn{1}{c|}{\textbf{MSE}} & \multicolumn{1}{c|}{\textbf{MAE}} & \multicolumn{1}{c|}{\textbf{MSE}} & \multicolumn{1}{c|}{\textbf{MAE}} \\
\hline
\textbf{NP} & \textbf{0.2377} & \textbf{0.2643} & 0.2529 & 0.2755 & \textbf{0.2294} & \textbf{0.2525} & 0.2498 & 0.2700 & \textbf{0.2300} & \textbf{0.2599} & 0.2468 & 0.2764 \\
\hline
\textbf{PJM} & \textbf{0.0837} & \textbf{0.1726} & 0.0853 & 0.1760 & \textbf{0.0706} & \textbf{0.1618} & 0.0753 & 0.1694 & \textbf{0.0781} & \textbf{0.1721} & 0.0940 & 0.1878 \\
\hline
\textbf{BE} & \textbf{0.3591} & \textbf{0.2425} & 0.3890 & 0.2521 & \textbf{0.3672} & \textbf{0.2516} & 0.3912 & 0.2673 & \textbf{0.3689} & \textbf{0.2519} & 0.3770 & 0.2569 \\
\hline
\textbf{FR} & \textbf{0.3593} & \textbf{0.1903} & 0.3784 & 0.1930 & \textbf{0.3582} & \textbf{0.2006} & 0.4273 & 0.2345 & \textbf{0.3700} & \textbf{0.1883} & 0.3804 & 0.2014 \\
\hline
\textbf{DE} & 0.4790 & 0.4256 & \textbf{0.4527} & \textbf{0.4220} & \textbf{0.4112} & \textbf{0.3981} & 0.5765 & 0.4804 & \textbf{0.4432} & \textbf{0.4179} & 0.4495 & 0.4203 \\
\hline
\textbf{Avg} & \textbf{0.3037} & \textbf{0.2591} & 0.3117 & 0.2637 & \textbf{0.2873} & \textbf{0.2529} & 0.3440 & 0.2843 & \textbf{0.2981} & \textbf{0.2580} & 0.3095 & 0.2686 \\
\hline
\end{tabular}
}
\end{center}
\end{table}

Note that the hyperparameters the original authors used varied per model and per market. Because there is no reason to assume the same hyperparameters will be optimal when Halo is applied, the Halo-modified models received a hyperparameter search of their own on the validation dataset, which fixed the hyperparameters used on the test datasets above. To see the effect of this, look at Table~\ref{tab:hyperparameters}, which includes the test results TimeXer with Halo would have given if it had kept the original authors' configuration instead of retuning. Note that in some cases the newly tuned parameters helped, but in others they did not, though in both the tuned and the untuned cases, Halo generally improved results. Running the hyperparameter search on the final test data would have made at least one of the headline numbers look better: the MSE on the DE market, though this would have been a form of ``statistical cheating''.

\begin{table}[h]
\caption{For MSE/MAE, the best numbers are shown in bold, and the second-best numbers are underlined. The left column runs TimeXer + Halo on the hyperparameter configuration TimeXer's authors found. The middle column retunes those hyperparameters for Halo. The right column runs TimeXer itself, on its authors' tuned hyperparameters.}
\label{tab:hyperparameters}
\begin{center}
\begin{tabular}{|l|r|r|r|r|r|r|}
\hline
 & \multicolumn{2}{c|}{\textbf{TimeXer + Halo, untuned}} & \multicolumn{2}{c|}{\textbf{TimeXer + Halo, tuned}} & \multicolumn{2}{c|}{\textbf{TimeXer}} \\
\cline{2-7}
 & \multicolumn{1}{c|}{\textbf{MSE}} & \multicolumn{1}{c|}{\textbf{MAE}} & \multicolumn{1}{c|}{\textbf{MSE}} & \multicolumn{1}{c|}{\textbf{MAE}} & \multicolumn{1}{c|}{\textbf{MSE}} & \multicolumn{1}{c|}{\textbf{MAE}} \\
\hline
\textbf{NP} & \textbf{0.2361} & \textbf{0.2621} & \underline{0.2377} & \underline{0.2643} & 0.2529 & 0.2755 \\
\hline
\textbf{PJM} & \textbf{0.0786} & \textbf{0.1721} & \underline{0.0837} & \underline{0.1726} & 0.0853 & 0.1760 \\
\hline
\textbf{BE} & \underline{0.3745} & \underline{0.2448} & \textbf{0.3591} & \textbf{0.2425} & 0.3890 & 0.2521 \\
\hline
\textbf{FR} & \underline{0.3710} & \underline{0.1921} & \textbf{0.3593} & \textbf{0.1903} & 0.3784 & 0.1930 \\
\hline
\textbf{DE} & \textbf{0.4514} & \underline{0.4243} & 0.4790 & 0.4256 & \underline{0.4527} & \textbf{0.4220} \\
\hline
\textbf{Avg} & \textbf{0.3023} & \underline{0.2591} & \underline{0.3037} & \textbf{0.2591} & 0.3117 & 0.2637 \\
\hline
\end{tabular}

\end{center}
\end{table}

Section~\ref{sec:from-point-estimates-to-heteroscedastic-estimates} describes the two different architectures for Halo: a parallel architecture for simple models and a dual-head architecture for more complex models. Hyperparameter tuning tested both architectures. For CrossLinear, on the validation set, the parallel architecture clearly outperformed the dual-head architecture, and that is the variant behind the headline numbers. Table~\ref{tab:crosslinear} includes test results with the dual-head architecture on CrossLinear. Here the difference between the two was less decisive than on the validation set.

\begin{table}[h]
\caption{Comparing CrossLinear with parallel Halo, dual-head Halo, and no Halo. For MSE/MAE, the best numbers are shown in bold, and the second-best numbers are underlined.}
\label{tab:crosslinear}
\begin{center}
\begin{tabular}{|l|r|r|r|r|r|r|}
\hline
 & \multicolumn{2}{c|}{\textbf{CrossLinear + parallel Halo}} & \multicolumn{2}{c|}{\textbf{CrossLinear + dual-head Halo}} & \multicolumn{2}{c|}{\textbf{CrossLinear}} \\
\cline{2-7}
 & \multicolumn{1}{c|}{\textbf{MSE}} & \multicolumn{1}{c|}{\textbf{MAE}} & \multicolumn{1}{c|}{\textbf{MSE}} & \multicolumn{1}{c|}{\textbf{MAE}} & \multicolumn{1}{c|}{\textbf{MSE}} & \multicolumn{1}{c|}{\textbf{MAE}} \\
\hline
\textbf{NP} & \underline{0.2300} & \textbf{0.2599} & \textbf{0.2298} & \underline{0.2626} & 0.2468 & 0.2764 \\
\hline
\textbf{PJM} & \underline{0.0781} & \textbf{0.1721} & \textbf{0.0780} & \underline{0.1741} & 0.0940 & 0.1878 \\
\hline
\textbf{BE} & \textbf{0.3689} & \textbf{0.2519} & 0.3936 & 0.2604 & \underline{0.3770} & \underline{0.2569} \\
\hline
\textbf{FR} & \underline{0.3700} & \textbf{0.1883} & \textbf{0.3520} & \underline{0.1897} & 0.3804 & 0.2014 \\
\hline
\textbf{DE} & \underline{0.4432} & \textbf{0.4179} & \textbf{0.4417} & 0.4220 & 0.4495 & \underline{0.4203} \\
\hline
\textbf{Avg} & \textbf{0.2981} & \textbf{0.2580} & \underline{0.2990} & \underline{0.2618} & 0.3095 & 0.2686 \\
\hline
\end{tabular}

\end{center}
\end{table}

For the other two models, the difference was less pronounced in hyperparameter tuning, so the less expensive dual-head architecture is the one behind those headline numbers. The same ``what-if'' scenario tested both Halo architectures on TimeXer. See Table~\ref{tab:timexer}.

\begin{table}[h]
\caption{Comparing TimeXer with parallel Halo, dual-head Halo, and no Halo. For MSE/MAE, the best numbers are shown in bold, and the second-best numbers are underlined.}
\label{tab:timexer}
\begin{center}
\begin{tabular}{|l|r|r|r|r|r|r|}
\hline
 & \multicolumn{2}{c|}{\textbf{TimeXer + parallel Halo}} & \multicolumn{2}{c|}{\textbf{TimeXer + dual-head Halo}} & \multicolumn{2}{c|}{\textbf{TimeXer}} \\
\cline{2-7}
 & \multicolumn{1}{c|}{\textbf{MSE}} & \multicolumn{1}{c|}{\textbf{MAE}} & \multicolumn{1}{c|}{\textbf{MSE}} & \multicolumn{1}{c|}{\textbf{MAE}} & \multicolumn{1}{c|}{\textbf{MSE}} & \multicolumn{1}{c|}{\textbf{MAE}} \\
\hline
\textbf{NP} & \textbf{0.2288} & \textbf{0.2589} & \underline{0.2377} & \underline{0.2643} & 0.2529 & 0.2755 \\
\hline
\textbf{PJM} & \underline{0.0842} & \underline{0.1726} & \textbf{0.0837} & \textbf{0.1726} & 0.0853 & 0.1760 \\
\hline
\textbf{BE} & \underline{0.3754} & \underline{0.2511} & \textbf{0.3591} & \textbf{0.2425} & 0.3890 & 0.2521 \\
\hline
\textbf{FR} & \underline{0.3597} & \textbf{0.1863} & \textbf{0.3593} & \underline{0.1903} & 0.3784 & 0.1930 \\
\hline
\textbf{DE} & \textbf{0.4510} & \underline{0.4248} & 0.4790 & 0.4256 & \underline{0.4527} & \textbf{0.4220} \\
\hline
\textbf{Avg} & \textbf{0.2998} & \textbf{0.2587} & \underline{0.3037} & \underline{0.2591} & 0.3117 & 0.2637 \\
\hline
\end{tabular}

\end{center}
\end{table}

\section{Conclusion}
\label{sec:conclusion}

This paper introduces Halo, which gives an existing DNN forecaster a second output for the scale parameter of its implied distribution and trains it under the matching negative log likelihood. The modification needs no architecture-specific design, and it improves the point estimate. Across three architectures --- a transformer, a GNN paired with a VAE, and a single-layer CNN --- and five electricity price markets, Halo improved MSE and MAE in 28 of 30 comparisons, and improved every model's average on both metrics. Average MSE fell between 2.6\% and 16.5\%, and average MAE between 1.7\% and 11.0\%, depending on the model.

Two practical findings came out of the experiments. First, how the scale estimate attaches to the network matters less than going from point estimates to heteroscedastic estimates in the first place. The two architectures described in Section~\ref{sec:from-point-estimates-to-heteroscedastic-estimates} finish within a percent and a half of each other on both models, and in the same direction each time: the parallel architecture is slightly ahead of the dual-head architecture on TimeXer, and slightly ahead again on CrossLinear, but never by enough to choose on accuracy alone. The dual-head architecture costs roughly half the parameters of the parallel one, so that is the one to prefer, with the parallel architecture worth testing as a variant when the budget allows. Second, the hyperparameters tuned for a point-estimate model are often sufficient for its Halo variant. Retuning on the validation set helped in two markets and hurt in three, on both metrics, and left the test averages within half a percent of where they started, so retuning is optional, not a prerequisite for the improvement.

The result sits against \citet{stirn2023}. Both findings can hold at once: electricity prices carry the time-varying volatility that gives a scale estimate something to learn, and the models in that study had no such signal to work from. Testing Halo on other benchmarks and horizons, and across seeds, is the next step.

\bibliographystyle{iclr2027_conference}
\bibliography{references}

@inproceedings{wang2024,
  author    = {Wang, Yuxuan and Wu, Haixu and Dong, Jiaxiang and Qin, Guo and
               Zhang, Haoran and Liu, Yong and Qiu, Yunzhong and Wang, Jianmin and
               Long, Mingsheng},
  title     = {{TimeXer}: Empowering transformers for time series forecasting
               with exogenous variables},
  booktitle = {Advances in Neural Information Processing Systems},
  volume    = {37},
  pages     = {469--498},
  year      = {2024},
}

@misc{borovykh2017,
  author       = {Borovykh, Anastasia and Bohte, Sander and Oosterlee, Cornelis W.},
  title        = {Conditional time series forecasting with convolutional neural
                  networks},
  howpublished = {arXiv:1703.04691},
  year         = {2017},
}

@book{box2015,
  author    = {Box, George E. P. and Jenkins, Gwilym M. and Reinsel, Gregory C.
               and Ljung, Greta M.},
  title     = {Time series analysis: forecasting and control},
  publisher = {John Wiley \& Sons},
  year      = {2015},
}

@book{chincarini2022,
  author    = {Chincarini, Ludwig B. and Kim, Daehwan},
  title     = {{Quantitative Equity Portfolio Management: An Active Approach to
               Portfolio Construction and Management}},
  edition   = {2nd},
  publisher = {McGraw Hill},
  year      = {2022},
}

@article{granger1969,
  author  = {Granger, Clive W. J.},
  title   = {Investigating causal relations by econometric models and
             cross-spectral methods},
  journal = {Econometrica},
  volume  = {37},
  number  = {3},
  pages   = {424--438},
  year    = {1969},
}

@article{hewamalage2023,
  author  = {Hewamalage, Hansika and Ackermann, Klaus and Bergmeir, Christoph},
  title   = {Forecast evaluation for data scientists: common pitfalls and best
             practices},
  journal = {Data Mining and Knowledge Discovery},
  volume  = {37},
  number  = {2},
  pages   = {788--832},
  year    = {2023},
}

@inproceedings{kim2021,
  author    = {Kim, Taesung and Kim, Jinhee and Tae, Yunwon and Park, Cheonbok
               and Choi, Jang-Ho and Choo, Jaegul},
  title     = {Reversible instance normalization for accurate time-series
               forecasting against distribution shift},
  booktitle = {International Conference on Learning Representations},
  year      = {2021},
}

@inproceedings{lai2018,
  author    = {Lai, Guokun and Chang, Wei-Cheng and Yang, Yiming and Liu,
               Hanxiao},
  title     = {Modeling long- and short-term temporal patterns with deep neural
               networks},
  booktitle = {The 41st International {ACM} {SIGIR} Conference on Research and
               Development in Information Retrieval},
  year      = {2018},
}

@techreport{lapedes1987,
  author      = {Lapedes, Alan and Farber, Robert},
  title       = {Nonlinear signal processing using neural networks: Prediction
                 and system modelling},
  institution = {Los Alamos National Laboratory},
  number      = {LA-UR-87-2662},
  year        = {1987},
}

@inproceedings{li2026,
  author    = {Li, Zhengyu and Qiu, Xiangfei and Zhu, Yuhan and Wu, Xingjian
               and Hu, Jilin and Guo, Chenjuan and Yang, Bin},
  title     = {{GCGNet}: Graph-consistent generative network for time series
               forecasting with exogenous variables},
  booktitle = {International Conference on Learning Representations},
  year      = {2026},
}

@inproceedings{liu2022,
  author    = {Liu, Yong and Wu, Haixu and Wang, Jianmin and Long, Mingsheng},
  title     = {Non-stationary transformers: Exploring the stationarity in time
               series forecasting},
  booktitle = {Advances in Neural Information Processing Systems},
  volume    = {35},
  pages     = {9881--9893},
  year      = {2022},
}

@misc{nie2022,
  author       = {Nie, Yuqi and Nguyen, Nam H. and Sinthong, Phanwadee and
                  Kalagnanam, Jayant},
  title        = {A time series is worth 64 words: Long-term forecasting with
                  transformers},
  howpublished = {arXiv:2211.14730},
  year         = {2022},
}

@inproceedings{nix1994,
  author    = {Nix, David A. and Weigend, Andreas S.},
  title     = {Estimating the mean and variance of the target probability
               distribution},
  booktitle = {Proceedings of 1994 {IEEE} International Conference on Neural
               Networks ({ICNN}'94)},
  volume    = {1},
  pages     = {55--60},
  publisher = {IEEE},
  year      = {1994},
}

@misc{oreshkin2019,
  author       = {Oreshkin, Boris N. and Carpov, Dmitri and Chapados, Nicolas
                  and Bengio, Yoshua},
  title        = {{N-BEATS}: Neural basis expansion analysis for interpretable
                  time series forecasting},
  howpublished = {arXiv:1905.10437},
  year         = {2019},
}

@inproceedings{qin2017,
  author    = {Qin, Yao and Song, Dongjin and Cheng, Haifeng and Cheng, Wei and
               Jiang, Guofei and Cottrell, Garrison W.},
  title     = {A dual-stage attention-based recurrent neural network for time
               series prediction},
  booktitle = {Proceedings of the 26th International Joint Conference on
               Artificial Intelligence},
  pages     = {2627--2633},
  publisher = {AAAI Press},
  year      = {2017},
}

@article{salinas2020,
  author  = {Salinas, David and Flunkert, Valentin and Gasthaus, Jan and
             Januschowski, Tim},
  title   = {{DeepAR}: Probabilistic forecasting with autoregressive recurrent
             networks},
  journal = {International Journal of Forecasting},
  volume  = {36},
  number  = {3},
  pages   = {1181--1191},
  year    = {2020},
}

@misc{seitzer2022,
  author       = {Seitzer, Maximilian and Tavakoli, Arash and Antic, Dimitrije
                  and Martius, Georg},
  title        = {On the pitfalls of heteroscedastic uncertainty estimation with
                  probabilistic neural networks},
  howpublished = {arXiv:2203.09168},
  year         = {2022},
}

@article{smyl2020,
  author  = {Smyl, Slawek},
  title   = {A hybrid method of exponential smoothing and recurrent neural
             networks for time series forecasting},
  journal = {International Journal of Forecasting},
  volume  = {36},
  number  = {1},
  pages   = {75--85},
  year    = {2020},
}

@inproceedings{stirn2023,
  author    = {Stirn, Andrew and Wessels, Hans-Hermann and Schertzer, Megan and
               Pereira, Laura and Sanjana, Neville E. and Knowles, David A.},
  title     = {Faithful heteroscedastic regression with neural networks},
  booktitle = {International Conference on Artificial Intelligence and
               Statistics},
  publisher = {PMLR},
  year      = {2023},
}

@inproceedings{vaswani2017,
  author    = {Vaswani, Ashish and Shazeer, Noam and Parmar, Niki and Uszkoreit,
               Jakob and Jones, Llion and Gomez, Aidan N. and Kaiser, {\L}ukasz
               and Polosukhin, Illia},
  title     = {Attention is all you need},
  booktitle = {Advances in Neural Information Processing Systems},
  volume    = {30},
  year      = {2017},
}

@article{werbos1988,
  author  = {Werbos, Paul J.},
  title   = {Generalization of backpropagation with application to a recurrent
             gas market model},
  journal = {Neural Networks},
  volume  = {1},
  number  = {4},
  pages   = {339--356},
  year    = {1988},
}

@inproceedings{wu2021,
  author    = {Wu, Haixu and Xu, Jiehui and Wang, Jianmin and Long, Mingsheng},
  title     = {{Autoformer}: Decomposition transformers with auto-correlation
               for long-term series forecasting},
  booktitle = {Advances in Neural Information Processing Systems},
  volume    = {34},
  pages     = {22419--22430},
  year      = {2021},
}

@inproceedings{zeng2023,
  author    = {Zeng, Ailing and Chen, Muxi and Zhang, Lei and Xu, Qiang},
  title     = {Are transformers effective for time series forecasting?},
  booktitle = {Proceedings of the {AAAI} Conference on Artificial Intelligence},
  volume    = {37},
  year      = {2023},
}

@inproceedings{zhou2021,
  author    = {Zhou, Haoyi and Zhang, Shanghang and Peng, Jieqi and Zhang, Shuai
               and Li, Jianxin and Xiong, Hui and Zhang, Wancai},
  title     = {{Informer}: Beyond efficient transformer for long sequence
               time-series forecasting},
  booktitle = {Proceedings of the {AAAI} Conference on Artificial Intelligence},
  volume    = {35},
  year      = {2021},
}

@inproceedings{zhou2022,
  author    = {Zhou, Tian and Ma, Ziqing and Wen, Qingsong and Wang, Xue and
               Sun, Liang and Jin, Rong},
  title     = {{FEDformer}: Frequency enhanced decomposed transformer for
               long-term series forecasting},
  booktitle = {International Conference on Machine Learning},
  publisher = {PMLR},
  year      = {2022},
}

@inproceedings{zhou2025,
  author    = {Zhou, Pengfei and Liu, Yunlong and Liang, Junli and Song, Qi
               and Li, Xiangyang},
  title     = {{CrossLinear}: Plug-and-play cross-correlation embedding for time
               series forecasting with exogenous variables},
  booktitle = {Proceedings of the 31st {ACM} {SIGKDD} Conference on Knowledge
               Discovery and Data Mining},
  year      = {2025},
}

@misc{qiu2025,
  author       = {Qiu, Xiangfei and Zhu, Yuhan and Li, Zhengyu and Wu, Xingjian
                  and Yang, Bin and Hu, Jilin},
  title        = {{DAG}: A dual correlation network for time series forecasting
                  with exogenous variables},
  howpublished = {arXiv:2509.14933},
  year         = {2025},
}

@article{lago2021,
  author  = {Lago, Jesus and Marcjasz, Grzegorz and De Schutter, Bart and
             Weron, Rafa{\l}},
  title   = {Forecasting day-ahead electricity prices: A review of
             state-of-the-art algorithms, best practices and an open-access
             benchmark},
  journal = {Applied Energy},
  volume  = {293},
  pages   = {116983},
  year    = {2021},
}

\appendix

\section{Implementation Details}
\label{sec:implementation-details}

Every model trains on a single Apple M2 Max system (30-core integrated GPU, 12 CPU cores, 32 GB unified memory) running macOS 26.6, where PyTorch 2.13 dispatches to the GPU through the Metal Performance Shaders (MPS) backend. The M2 Max shares one pool of memory between CPU and GPU, so the 32 GB figure is the whole budget for a run rather than a dedicated accelerator memory. Training runs for up to 50 epochs, with an early stopping patience of 5. That budget matches GCGNet, whose authors trained for more epochs than TimeXer and CrossLinear, and it puts all three models on roughly equal footing. The optimizer is Adam, with the initial learning rate left as a tunable hyperparameter. That rate holds for the first four epochs, then falls to 0.9 of its previous value at each epoch after that. Every reported number is a single training run. A fixed base seed (2021) combines with the model and the market to give each run its own seed, so no result is an average over restarts, and the paper reports no variance across seeds. All hyperparameter tuning uses a validation hold-out set, and every reported result comes from a separate test hold-out set, touched only after tuning finished. Since this is time series data, the training data comes chronologically first, then the validation data, then the test data.

\end{document}